%% file: 00-main.tex
\documentclass[10pt,twocolumn,letterpaper]{article}

\usepackage[pagenumbers]{cvpr} 

\usepackage{graphicx}
\usepackage{amsmath}
\usepackage{amssymb}
\usepackage{booktabs}

\usepackage[pagebackref,breaklinks,colorlinks]{hyperref}

\usepackage{pdfcomment}

\usepackage[capitalize]{cleveref}
\crefname{section}{Sec.}{Secs.}
\Crefname{section}{Section}{Sections}
\Crefname{table}{Table}{Tables}
\crefname{table}{Tab.}{Tabs.}

\def\cvprPaperID{*****} 
\def\confName{CVPR}
\def\confYear{2022}

\begin{document}

\title{Towards Global-Scale Crowd+AI Techniques\linebreak 
to Map and Assess Sidewalks for People with Disabilities}

\author{
\textbf{Maryam Hosseini$^{1,2}$, Mikey Saugstad$^{3}$, Fabio Miranda$^{4}$,}\\
\textbf{Andres Sevtsuk$^{5}$, Claudio T. Silva$^{2}$, Jon E. Froehlich$^{3}$}\\
\fontsize{10pt}{10pt}\selectfont $^{1}$~Rutgers University, $^{2}$~NYU, $^{3}$~University of Washington, $^{4}$~University of Illinois at Chicago, $^{5}$~MIT
}

\maketitle

\begin{abstract}
There is a lack of data on the location, condition, and accessibility of sidewalks across the world, which not only impacts where and how people travel but also fundamentally limits interactive mapping tools and urban analytics. In this paper, we describe initial work in semi-automatically building a sidewalk network topology from satellite imagery using hierarchical multi-scale attention models, inferring surface materials from street-level images using active learning-based semantic segmentation, and assessing sidewalk condition and accessibility features using Crowd+AI. We close with a call to create a database of labeled satellite and streetscape scenes for sidewalks and sidewalk accessibility issues along with standardized benchmarks. 
\end{abstract}


\input{01-intro}
\input{02-1-sidewalk-extraction}

\input{02-2-sidewalk-topology}
\input{02-3-surface-material}
\input{02-4-project-sidewalk}
\input{03-poc}
\input{04-discussion}

{\small
\bibliographystyle{ieee_fullname}
\bibliography{paper}
}

\end{document}

%% file: 01-intro.tex
\section{Introduction}


Sidewalks form the backbone of cities. At their best, they offer sustainable transit, help interconnect mass transportation services, and support local commerce and recreation. For people with disabilities, sidewalks support independence, physical activity, and overall quality of life \cite{Christensen2010, Eisenberg2017, Harris2015, Mitchell2006}. Despite decades of civil rights legislation, however, city streets and sidewalks remain inaccessible\cite{ADA1990}. As the UN notes “\textit{[there is a] widespread lack of accessibility in built environments, from roads and housing to public buildings and spaces}” \cite{UNNewUrbanAgenda2020}.

The problem is not just a lack of accessible sidewalks but also a lack of reliable data on where sidewalks exist and their quality\cite{Eisenberg2020, deitz2021free,Froehlich2019}. In a sample of 178 US cities, Deitz \textit{et al.} found that only 36 (20\%) published sidewalk data, 18 (10\%) had curb ramp locations, and even fewer included detailed accessibility information like sidewalk condition, obstructions, and cross controls~\cite{DeitzSqueakyWheels2021}. This lack of data fundamentally limits how sidewalks can be studied in cities, the ability for communities, disability advocacy groups, and local governments to understand, transparently discuss, and make informed urban planning decisions, and how sidewalks and accessibility are incorporated into interactive map, navigation, and GIS tools\cite{Froehlich2019,10.1145/3313831.3376399}.

We argue that any comprehensive analysis of pedestrian infrastructure must include a threefold understanding of \textit{where} sidewalks are, \textit{how} they are connected, and \textit{what} their condition is. In this paper, we introduce an initial semi-automatic pipeline that maps \textit{sidewalk locations}, infers \textit{surface materials}, and applies an \textit{accessibility rating} using a combination of crowdsourcing and machine learning techniques (\cref{fig:pipeline}). To demonstrate its potential, we apply our pipeline to Washington DC and create different visualizations of sidewalk connectivity and accessibility. We close with a discussion of key areas of open research that intersect computer vision, HCI, accessibility, and urban informatics. 


%% file: 02-1-sidewalk-extraction.tex
\section{Crowd+AI Sidewalk Pipeline}
\label{network}

At the core of our contribution is the threefold integration of sidewalk \textit{location}, \textit{connectivity}, and \textit{condition}. All three are critical to assessing pedestrian infrastructure and building pedestrian-oriented routing tools. To achieve this, we propose a four-stage \emph{Crowd+AI Sidewalk Pipeline} that leverages vision and crowdsourced techniques as well as aerial and street-level imagery to enable network-level sidewalk assessments. We describe each pipeline stage below.


\subsection{Extracting Sidewalks from Aerial Imagery}



\begin{figure}[t]
  \centering
  \pdftooltip{\includegraphics[width=\linewidth]{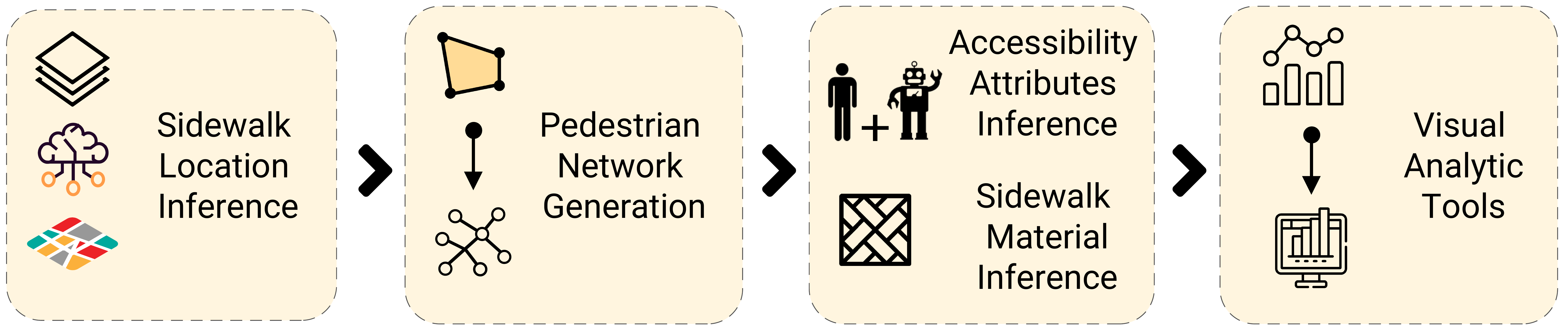}}{Image showing the stages of the Crowd+AI sidewalk pipeline.}
  \vspace{-0.5cm}
  \caption{We introduce a four-stage Crowd+AI sidewalk pipeline that combines computer vision and crowdsourcing to \textit{locate} sidewalks, build a \textit{network topology}, infer \textit{surface material}, and \textit{assess accessibility}. The resulting output can be used to support accessibility-aware pedestrian routing and new urban science analyses centered on equity and access.}
  \vspace{-0.6cm}
  \label{fig:pipeline}
\end{figure}

Our pipeline begins with the extraction of \textit{pedestrian pathways}—including sidewalks, footpaths, and crosswalks—from aerial imagery using semantic segmentation. Although semantic segmentation has been broadly used to detect roads and buildings from aerial images \cite{balali2015detection, iglovikov2017satellite, li2019semantic} and to auto-generate road network topologies~\cite{bastani2018roadtracer, wei2019road, etten2020city}, it has not been widely applied to pedestrian infrastructure—perhaps due to  two key challenges. First, semantic segmentation algorithms require large-scale, high-quality training datasets, which can be labor-intensive and costly to prepare. Thus, researchers often rely on pre-existing publicly available models pre-trained on datasets such as \textit{CityScapes}~\cite{cordts2016cityscapes}, \textit{Mapillary}~\cite{neuhold2017mapillary}, and \textit{ADE20K}~\cite{zhou2017scene}, which historically underemphasize pedestrian-related features. Second, compared to roads and buildings, detecting sidewalks, footpaths, and crosswalks is more challenging due to their comparatively smaller visual footprints and occlusion from shadows, vegetation, and tall structures~\cite{hosseini2021sidewalk}.

\begin{figure}[t]
  \centering
  \pdftooltip{\includegraphics[width=1\linewidth]{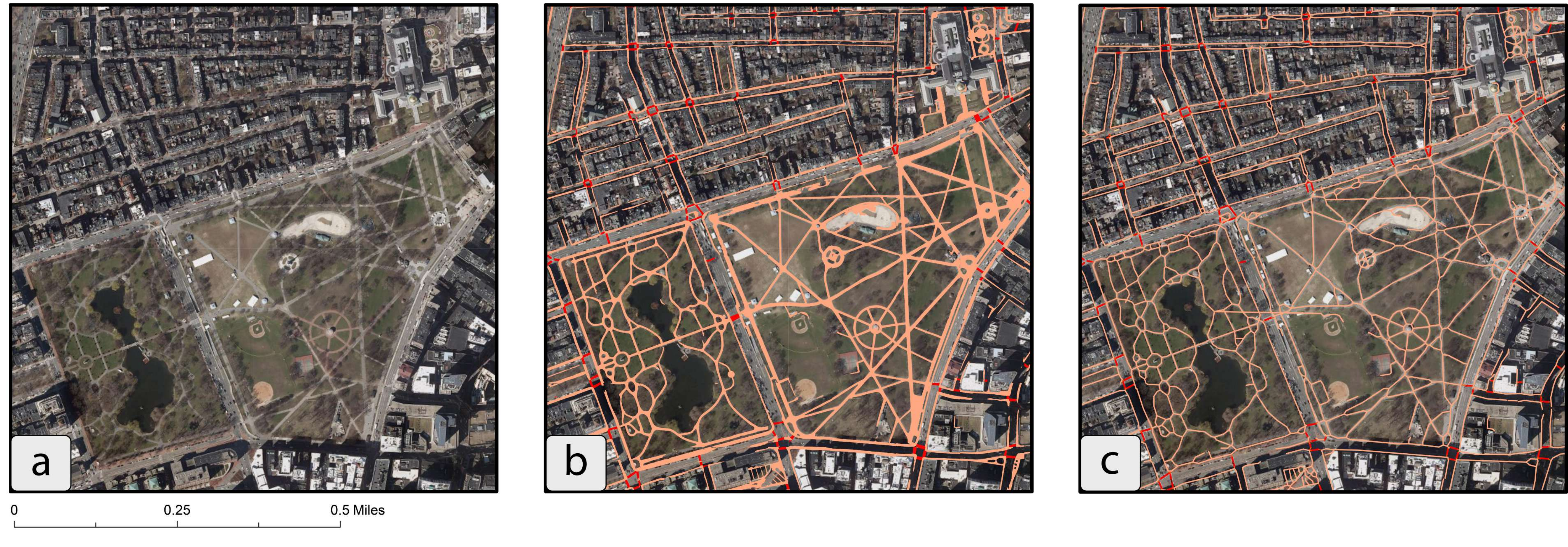}}{Aerial images highlighting the extraction of labeled sidewalk data and the creation of a pedestrian network.}
  \vspace{-0.5cm}
  \caption{In Stage 1, we input aerial imagery (a) to extract labeled sidewalk raster (b) that are used in Stage 2 (c) to create a pedestrian network~\cite{hosseini2022network}.}
  \vspace{-0.6cm}
  \label{fig:stage1}
\end{figure}

To detect pedestrian infrastructure, we first feed publicly available orthorectified aerial image tiles into a semantic segmentation model (\cref{fig:stage1}), which outputs a labeled raster image. Each pixel is labeled with one of four classes: \textit{sidewalks} (including footpaths), \textit{crosswalks}, \textit{roads}, and \textit{background}. To address segmentation challenges, we use a hierarchical multi-scale attention model~\cite{tao2020hierarchical}—the attention mechanism allows the model to focus on the most relevant features as needed~\cite{chen2016attention}. For our backbone, we use a composite \textit{HRNet-W48+OCR} model—one of the top-performing models across multiple vision benchmarks \cite{cordts2016cityscapes} and performs well in detecting fine-level details~\cite{wang2020deep}. For example, our approach can distinguish between visually similar classes such as asphalt roads and sidewalks.

For training and testing, we use pre-existing open-government datasets drawn from three US cities: Cambridge, MA; Washington DC; and New York City. While the availability of such data is generally limited ~\cite{deitz2021free}, these pre-existing datasets allowed us to bypass laborious manual labeling. Still, challenges remained. To create a clean dataset, we accommodated different mapping standards across municipalities and addressed temporal differences between the GIS data and the aerial images. In total, our research team manually corrected 2,500 image tiles of 12,000 in the training set (20.8\%), 1,620 of 4,000 in validation (40.5\%), and 1,500 of 4,000 in test (37.5\%).

To evaluate performance, we split our data into 60\% training, 20\% validation, and 20\% test. We then trained our model over 310 epochs using four NVIDIA RTX~8000 GPUs with 48 GB of RAM and a batch size of 16, a stochastic gradient descent for the optimizer with a polynomial learning rate~\cite{liu2015parsenet}, momentum 0.9, weight decay $5e^{-4}$, and an initial learning rate of 0.002. To handle class imbalance, we employed uniform sampling in the data loader~\cite{zhu2019improving}. For our primary performance metric, we use the Jaccard index~\cite{jaccard1908nouvelles}—commonly referred to as the \textit{intersection over union} (IoU) approach, which is a scale-invariant standard evaluation metric for semantic segmentation tasks. We do not compute pixel-level accuracy since sidewalk features comprise a small portion of each tile, resulting in a significant class imbalance and artificially high accuracy. On average, our model achieved 84.5\% IoU  accuracy with road detection performing best (86.0\%) followed by sidewalks (82.7\%) and crosswalks (75.4\%). In a qualitative review of results, we found that incorrect crosswalk detection were due to a lack of clear visual demarcations and faded paint. 

%% file: 02-2-sidewalk-topology.tex
\subsection{Creating Sidewalk Network Topologies}
The Stage 1 detection model outputs labeled pixels in rasterized format, which is fed into our Stage 2 pedestrian network creation algorithm. This algorithm has two key parts: first, we convert the labeled rasters to georeferenced polygons using connected-component labeling\cite{rosenfeld1966sequential, he2009fast}—which finds contiguous pixel groups within the same class to form regions or raster polygons. We then map these polygonal elements into geographic coordinates. Second, to create a node-network diagram of sidewalk connectivity, we use computational geometry techniques to convert the polygons into polylines (the centerline of the polygon). 

To evaluate performance of the polygon extractor, we compute the mean aerial overlap percentage between our extracted polygons and those in the official city GIS datasets. Overall, the percentage overlap in Cambridge was 98.9\%, NYC (98.3\%), and DC (84.4\%).
Finally, to evaluate the accuracy of our sidewalk networks, we compared our centerlines to official government datasets or, in the case of DC, \textit{OpenStreetMap} sidewalk data (as DC does not publish sidewalk topologies). Our model matched 83.1\% of all segments in Cambridge, 85.1\% of official footpath segments in New York, and 76.9\% of OSM sidewalk networks in DC. 

Overall, these results are promising and demonstrate the potential of automatically creating pedestrian networks from aerial imagery but also suggest opportunities for crowdsourced review and refinement.

%% file: 02-3-surface-material.tex
\begin{figure}[t]
  \centering
  \pdftooltip{\includegraphics[width=1\linewidth]{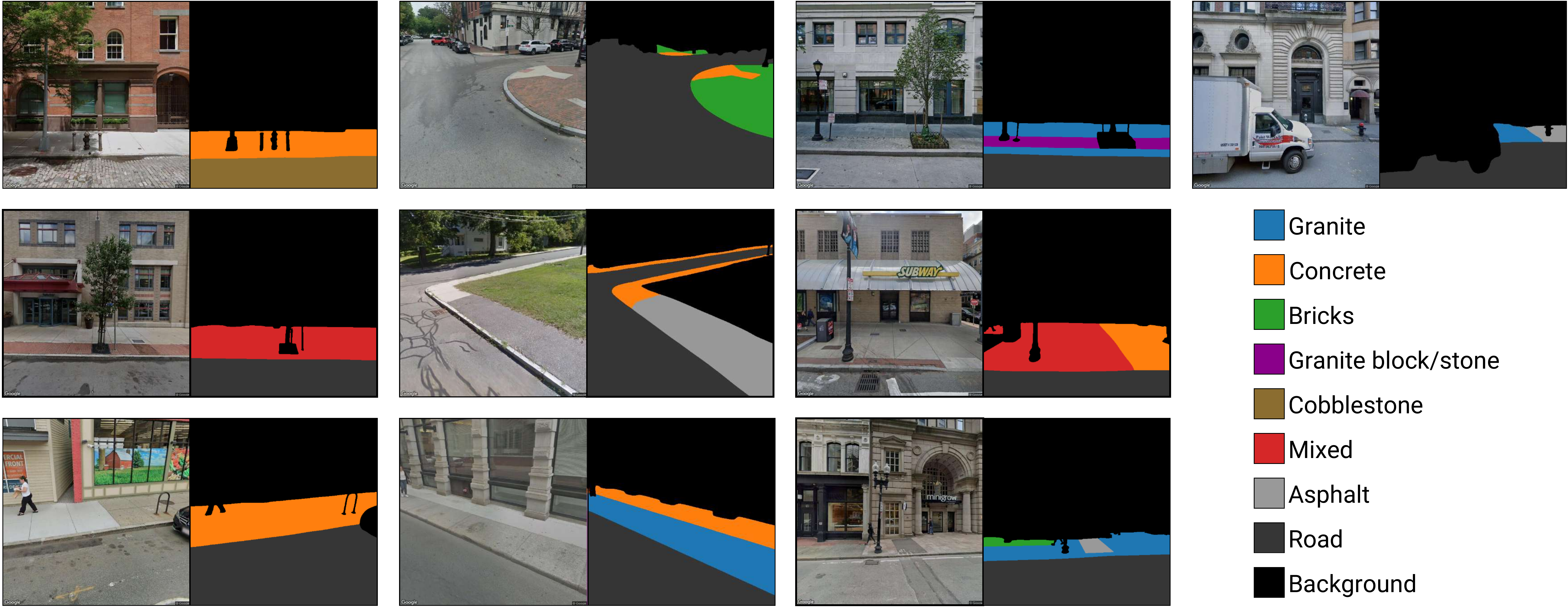}}{Images showing the segmentation of sidewalks according to their material: granite, concrete, bricks, granite block stone, cobblestone, mixed and asphalt.}
  \vspace{-0.5cm}
  \caption{Stage 3 auto-segments sidewalks from streetscape images and classifies eight surface materials, including concrete and brick ~\cite{hosseini2022citysurfaces}.}
  \vspace{-0.6cm}
  \label{fig:surfacematerial}
\end{figure}

\subsection{Inferring Sidewalk Surface Material}
\label{materials}
While Stages 1 and 2 produce a sidewalk network topology, they do not include an assessment of sidewalk surface composition (Stage 3) or its accessibility (Stage 4). Thus, in Stage 3, we examine techniques to automatically infer sidewalk surface materials (\cref{fig:surfacematerial}), such as concrete, brick, and cobblestone, which can have varying impacts on pedestrian safety and accessibility \cite{talbot2005falls, clifton2005pedestrian, AGHAABBASI2018475, chippendale2015}. We introduce \textit{CitySurfaces}~\cite{hosseini2022citysurfaces}, a deep-learning based framework that automatically classifies sidewalk materials using omnidirectional streetscape imagery—specifically, \textit{Google Street View} (GSV).  Because no pre-existing annotated image dataset exists for sidewalk materials, we pursued a three-phase active learning approach. 

\textbf{Phase 1.} To start the training process, we randomly sampled 1,000 streetscape images from Boston, MA, fed our sample into HRNet-W48~\cite{wang2020deep} pre-trained on the Cityscapes dataset ~\cite{cordts2016cityscapes}, and obtained initial segmentation results. While HRNet outputs 19 classes including \textit{sky}, \textit{trees}, and \textit{buildings}, we filter only to \textit{roads} and \textit{sidewalks}. To generate an initial set of labeled surface material data, we use the \textit{Boston Sidewalk Inventory}~\cite{Bostonswinv}—a unique open dataset that describes the dominant surface material of each sidewalk segment collected via manual field surveys: \textit{concrete}, \textit{brick}, \textit{granite}, \textit{concrete/brick mix}, and \textit{asphalt}.

\textbf{Phase 2.} We iteratively train an attention-based model using the labeled images from Phase 1. We begin with 800 images for training and 200 for validation with a batch size of 8 and similar hyperparameters to Stage 1. We train the model in multiple stages. At each stage (10 epochs), we choose the epoch with the highest average IoU on the validation set and qualitatively analyze results to guide new training data sampling. The weights from best epoch are used to initialize the next stage’s model with more training data. To sample new images, we examine the model’s uncertainty estimates and select images that performed worst. 
Following this sampling strategy, we retrieve 300 unlabeled images, apply the current model, correct the predicted labels and add them to the overall training set. To improve model generalization, we begin to include streetscape images from a second city: Manhattan. 
Manhattan images are fed to the model, the predicted results corrected, and added to the training dataset. 
We iterated this procedure for five stages until improvements waned. At the final stage, the model was trained on 2,500 images and achieved an average of 88.6\% on the held-out test set.

\textbf{Finally, for Phase 3,} we add three additional surface classes: \textit{granite blocks}, \textit{hexagonal pavers}, and \textit{cobblestone}. We manually annotate an additional 1,150 images that contain these new classes (800 for training, 150 for validation, 200 for test). The newly generated labeled set was used to train the model with Phase 2 weights and replaces the final softmax layer to produce ten output classes (8 surface materials plus \textit{road},  \textit{background}). At the end of each stage, we select a new sample of unlabeled images similar to Phase 2, obtain segmentation predictions, refine the results, and retrain the model. In total, 726 additional images were added to the training set. In the final stage, we used 3,226 training images (2,500 from Phase 2 and 726 from Phase 3). 

\textbf{Results.} To evaluate performance, we again avoid pixel-level accuracy since sidewalks comprise a relatively small portion of each streetscape. Overall, our model achieves 88.4\% mean IoU accuracy with \textit{hexagonal asphalt pavers} performing best (92.8\%) followed by \textit{asphalt} (92.6\%), \textit{brick} (91.8\%), and \textit{concrete} (88.7\%). This work demonstrates the potential of active learning in accurately identifying sidewalk surface materials in streetscape imagery.

%% file: 02-4-project-sidewalk.tex
\subsection{Crowd+AI Accessibility Assessments}
\label{ps}
The above stages produce sidewalk topologies and surface classifications—both which impact human mobility and people with disabilities—but neither focus specifically on \textit{accessibility}. Thus, in Stage 4, we introduce Crowd+AI techniques to semi-automatically find, label, and assess sidewalk accessibility features in the built environment such as \textit{curb ramps}, \textit{surface problems}, and \textit{obstacles}. In previous work, we demonstrated that online streetscape imagery is an accurate source for assessing accessibility infrastructure~\cite{Hara2013BusStop, Hara2015TACCESS} and that with our custom labeling tools, minimally trained crowdworkers could accurately and quickly find street-level accessibility problems~\cite{Hara2012ASSETSPoster, Hara2013CHI, Hara2015TACCESS}. However, relying solely on human labor limits scalability. We then explored how to effectively combine automated methods with crowd work ~\cite{Hara2013HCIC, Hara2013AAAI, hara2014tohme}. Our first hybrid Crowd+AI system, \textit{Tohme}, infers the difficulty of a sidewalk scene using a trained SVM and allocates work accordingly to either a computer vision-based pipeline or human users~\cite{hara2014tohme}. In a study of ~1,000 street intersections across four North American cities, Tohme performed similarly to a purely human labeling approach but was more efficient. While promising, Tohme was limited to a small training dataset and only supported one sidewalk feature (curb ramp recognition).

\begin{figure}[t]
  \centering
  \pdftooltip{\includegraphics[width=0.95\linewidth]{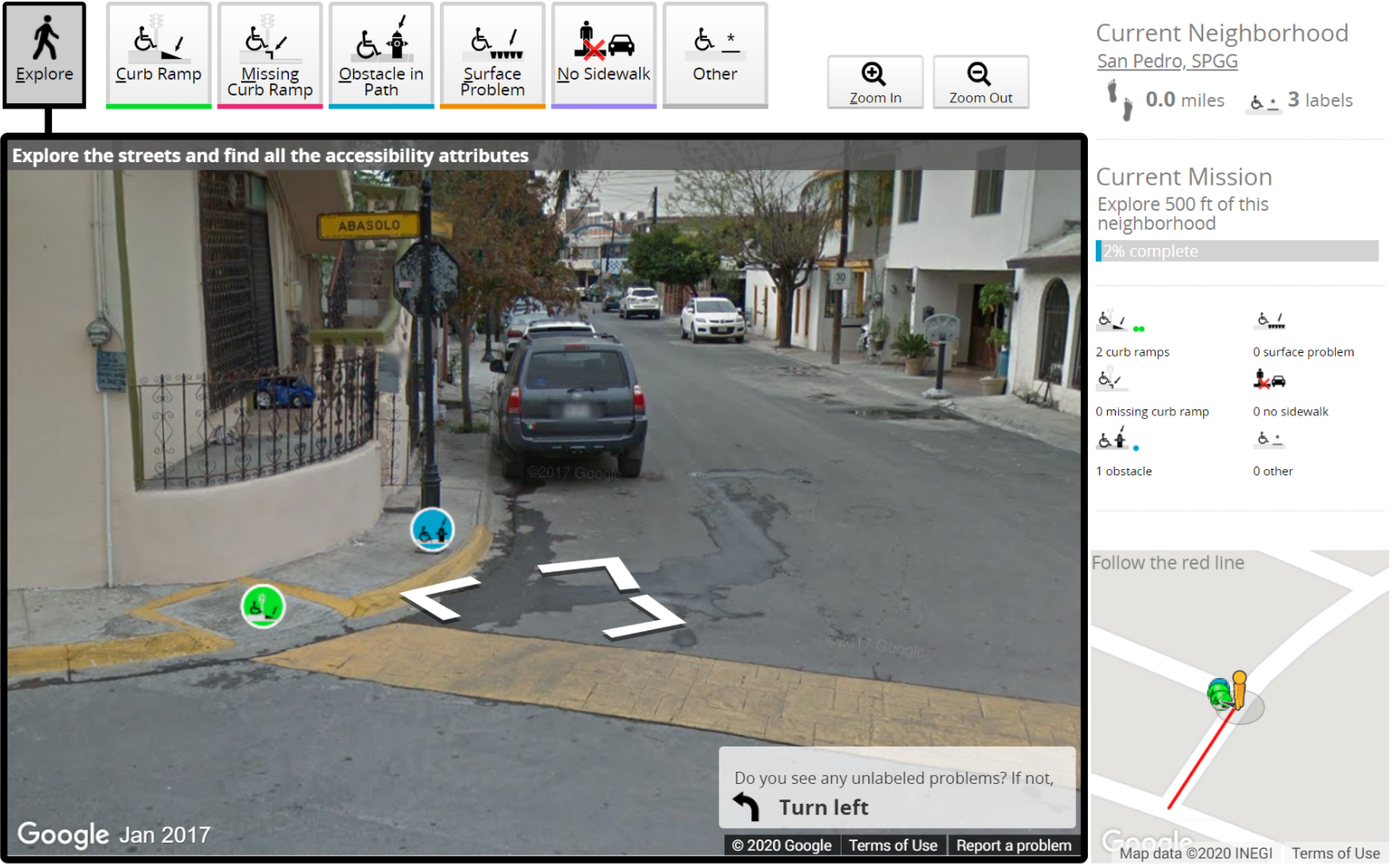}}{A screenshot of Project Sidewalk that shows an image of an intersection with a curb ramp labeled with a green "curb ramp" point and a light pole labeled with a blue "obstacle in path" point}
  \vspace{-0.25cm}
  \caption{Stage 4 uses Crowd+AI techniques to label accessibility features/barriers in the pedestrian environment. Above, a user labeled a \textit{curb ramp} (in green) and an \textit{obstacle} (in blue) in Project Sidewalk \cite{Saha2019ProjectSidewalk}}
  \vspace{-0.5cm}
  \label{fig:projectsidewalk}
\end{figure}

Thus, we began to explore more scalable approaches, culminating in \href{http://projectsidewalk.org}{\textit{Project Sidewalk}}—an interactive online tool that allows anyone with a laptop and Internet connection to remotely label accessibility problems by virtually walking through city streets in GSV, similar to a first-person video game (\cref{fig:projectsidewalk}). In a 2018 pilot deployment, 1,400 users virtually audited 2,934 km of  DC streets, providing 250,000 sidewalk accessibility labels ~\cite{Saha2019ProjectSidewalk}. With simple quality control mechanisms, we found that remote users could find and label 92\% of accessibility problems, including \textit{missing curb ramps}, \textit{obstacles}, \textit{surface problems}, and \textit{missing sidewalks}. To qualitatively assess reactions to our tool, we also conducted a complementary interview study with three stakeholder groups (\textit{N=}14)—government officials, people with disabilities, and caregivers. All felt that Project Sidewalk enabled rapid data collection, allowed for gathering diverse perspectives about accessibility, and helped engage citizens in urban design. Key concerns included data reliability and quality, which are ongoing research foci in our group.

Building on this DC pilot and working closely with local government partners and NGOs, we have now deployed Project Sidewalk in ten additional cities, including in Mexico and the Netherlands. Thus far, we have collected over 700,000 geo-located sidewalk accessibility labels and 400,000 validations—to our knowledge, the largest and most granular open sidewalk accessibility dataset ever collected. This large, ever-growing labeled dataset of images paired with advances in computer vision have enabled new deep learning methods for automatic sidewalk assessment. In Weld \textit{et al. }\cite{Weld2019DeepLearning}, we showed how a modified version of \textit{ResNet-18}—which incorporates LIDAR depth, scene position, and geography features in addition to pixels—could achieve state-of-the-art performance in automatically validating human  labels (average precision/recall: 81.3\%, 77.2\%). We also presented the first examination of cross-city model generalization showing that one city’s labels (DC) could be used to pre-train model weights for two other test cities (Seattle, WA and Newberg, OR). 

%% file: 03-poc.tex
\section{Demonstrating Proof-of-Concept}
To demonstrate the potential of our approach, we apply our four-stage pipeline to Washington DC and create sidewalk visualizations of topology, surface material, and accessibility. DC provides an interesting testbed: it has over 1,100 miles of city streets, diverse and historic urban designs, and is a popular tourist destination; however, no official pedestrian network data exists for the city. 

To extract the pedestrian pathways, we fed 73,000 orthorectified satellite image tiles obtained from \href{https://opendata.dc.gov/}{opendata.dc.gov} into our Stage 1 algorithm. Then, in Stage 2, we converted the auto-labeled sidewalk, footpath, and crosswalk rasters into georeferenced polygons and centerlines. Finally, to compute accessibility metrics, we incorporated both surface material inference data (from Stage 3) as well as crowdsourced accessibility information (from Stage 4). 

\begin{figure}[t]
  \centering
  \pdftooltip{\includegraphics[width=1\linewidth]{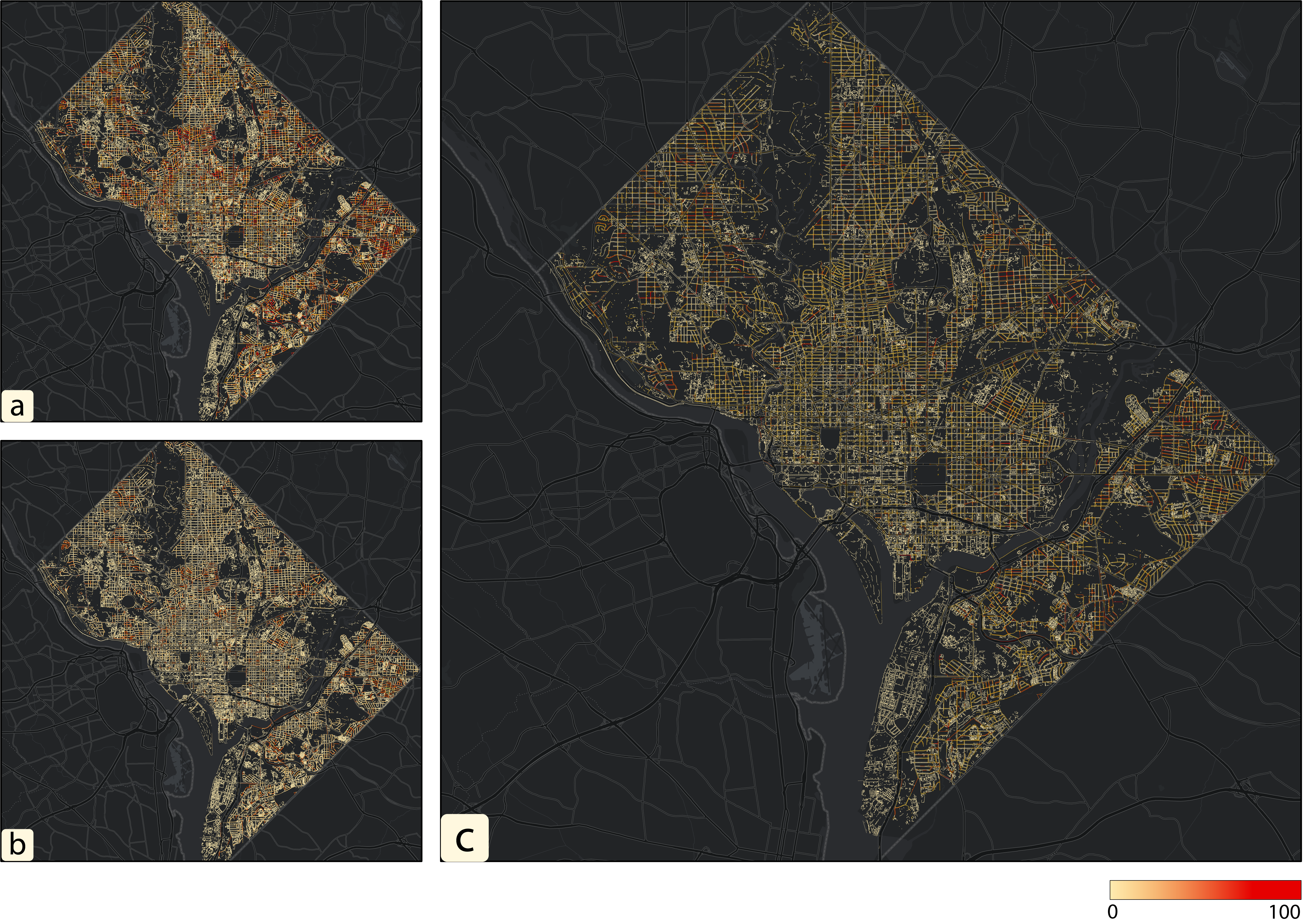}}{Image showing the proof-of-concept visualizations in Washington DC.}
  \vspace{-0.6cm}
  \caption{Proof-of-concept of our pipeline in Washington DC.}
  \vspace{-0.6cm}
  \label{fig:dc}
\end{figure}

We produce three proof-of-concept visualizations, which are based on computed sidewalk-accessibility scores—an open area of research \cite{LiASSETS18InteractiveModeling, Saha2022CHIUrbanVisualization}. First, we created a sidewalk heatmap visualization using Stage 4 accessibility data (\cref{fig:dc}a; red is worse). We differentiate between street crossings, which connect sidewalk segments, and the sidewalk segments themselves. For the street crossings, we associate \textit{curb ramps} and \textit{missing curb ramps} with intersections and compute a \textit{crossability score}. For the sidewalk segments, we calculate a severity-weighted sum of all accessibility problems over each sidewalk segment.  Second, we created a sidewalk heatmap visualization that incorporates Stage 3's surface material inference data (\cref{fig:dc}b). Here, we apply higher weights to bricks and cobblestone surfaces, which pose higher tripping hazards to people using mobility aids and bumpy, uncomfortable surfaces for wheelchair users. Finally, we created a hybrid visualization that incorporates both surface material (Stage 3) and accessibility (Stage 4) shown in \cref{fig:dc}c.

%% file: 04-discussion.tex
\section{Discussion and Future Work}
Our overarching vision is to develop scalable Crowd+AI techniques capable of \textit{mapping} and \textit{assessing} every sidewalk in the world. In this paper, we introduced a preliminary four-stage pipeline that extracts sidewalk \textit{locations}, infers \textit{surface materials}, and applies an \textit{accessibility} rating using a combination of computer vision and crowdsourcing. While prior work has examined each in piecemeal, we offer the first comprehensive pipeline towards addressing the grand challenge   of identifying \textit{where} sidewalks are, \textit{how} they are connected, and \textit{what} their condition is \cite{Froehlich2019}. All three are needed to create accessibility-aware pedestrian routing algorithms, interactive maps of neighborhood accessibility, and to enable equity analyses examining  sidewalk infrastructure availability/condition and key correlates such as race, real-estate pricing, and socio-economics.

Towards future work, we would like to examine: (1) how the crowd and AI can work together in each stage to improve efficiency and accuracy; (2) how our methods perform across varying urban fabrics and geographic contexts; (3) and advance understanding of the underlying biases in our methods—where do they fail and why?

Finally, we call on this cross-disciplinary community to create a database of high-quality, labeled satellite and streetscape scenes for sidewalks and sidewalk accessibility problems along with computer vision benchmarks, which has been so critical to innovation in other ML-based areas.  

\section*{Acknowledgments}
This work was funded in part by the National Science Foundation under grants SCC-IRG 2125087, CNS-1229185, CCF-1533564, CNS-1544753, CNS-1730396, CNS-1828576, CNS-1626098; the Pacific Northwest Transportation Consortium (PacTrans); UW CREATE; C2Smart; the Moore-Sloan Data Science Environment at NYU; and the NVIDIA NVAIL at NYU.
%
